%% file: main-uw.tex
\documentclass[10pt,logo]{uwtechreport}
\newcommand{\drop}[1]{%
  $\textcolor{BrickRed}{_{\scriptstyle -#1}}$%
}

\newcommand{\nodrop}{%
  $\textcolor{gray}{_{\scriptstyle \pm0.00}}$%
}

\usepackage{wrapfig}
\usepackage{array,natbib}

\usepackage{placeins,flafter}

\usepackage{multirow}

\graphicspath{{figs/}{assets/}}
\newcommand{\norm}[1]{\left\lVert#1\right\rVert}
\newcommand{\gain}[1]{%
  $\textcolor{ForestGreen}{_{\scriptstyle +#1}}$%
}

\usepackage{algorithm}
\usepackage{algpseudocode}

\newcommand{\ourwork}{\texttt{TAPDreamer}}

\title{TAPDreamer: Transferable Adversarial Patches for World Action Models}

\author{
  Xuanyu Lu$^{1\dagger}$ \; Fengqing Jiang$^{1\dagger\ddagger}$ \; Kaiyuan Zheng$^{1\dagger}$ \; Yichen Feng$^{1}$ \; Yaorui Ding$^{1}$ \; Yuetai Li$^{1}$ \; \newline
  Zhen Xiang$^{2}$ \; Bhaskar Ramasubramanian$^{1,3}$ \; Basel Alomair$^{1,4,5}$ \; Luyao Niu$^{1}$ \; Radha Poovendran$^{1}$ \\
  \small $^{1}$University of Washington \quad
         $^{2}$University of Georgia \quad
         $^{3}$Western Washington University \\
  \small $^{4}$King Abdulaziz City for Science and Technology \quad
         $^{5}$HUMAIN \\
  \small $^{\dagger}$Equal Contribution \quad $^{\ddagger}$Project Lead \\
  {\small \href{https://tapdreamer.github.io/}{\texttt{tapdreamer.github.io}}}
}
\begin{document}
\maketitle
\begin{abstract}
World models learn to predict how their environment will evolve, making them an important foundation for general-purpose robotic control. Yet World Action Models (WAMs) rely on camera inputs, and manipulating those inputs can corrupt the visual representations that underpin their tasks and action policies. 
Existing attacks on these models optimize against actions or predicted futures and require access to victim outputs or generation modules. We propose \ourwork{}, a transferable fixed local patch whose pixel optimization uses only a public visual encoder and requires no target-policy queries. Its design addresses two requirements: a local modification should affect representations beyond its footprint, and the resulting disturbance should persist across observations. We optimize a fixed local patch on six frames from one source task by maximizing the mean L1 distance between clean and patched encoder outputs over all latent positions and channels. Mechanism analysis reveals that interactions between patch-induced changes in attention weights and value vectors broadcast a nearly identical representation shift beyond the patch footprint, while a shared perturbation response persists across task observations despite differences in clean latents. In closed-loop evaluation, one frozen patch per benchmark, covering about 6.5\% of the input, reduces FastWAM's success rate from 97.65\% to 0.00\% across 40 LIBERO tasks and from 90.86\% to 0.00\% across 50 RoboTwin tasks; matched random patches retain 81.45\% and 79.20\% success. The same patches reduce success to 1.45\% and 1.00\% on two DreamWAM configurations and to 10.60\% on Motus. Behavioral analysis connects these representation shifts to grasping failures and separates rapid command recovery after patch removal from persistent physical consequences. These results show that protecting downstream action generation alone is insufficient: WAM defenses must also secure shared visual encoders against persistent local perturbations.
\end{abstract}
\abscontent

\input{sec/intro}

\input{sec/setup}

\input{sec/experiment}

\input{sec/mechanism}

\input{sec/related}
\input{sec/conclusion}

\subsection*{Ethics statement}
This study aims to identify vulnerabilities of World Action Models to local visual perturbations to inform robot safety evaluation and defense research. All experiments were conducted in the LIBERO and RoboTwin simulation environments, with attacks implemented by modifying model input images. We recognize that such methods could be misused to disrupt normal robot operation and emphasize that their use should be restricted to authorized security testing and research.
Our mechanism analysis suggests potential defenses that attenuate keys and values from suspicious image regions or train the encoder to preserve representations outside locally perturbed regions.
These interventions target the attention-mediated spread of the patch-induced shift; their effectiveness against adaptive attacks and their impact on clean-task performance remain to be evaluated.

\bibliographystyle{iclr2027_conference}
\bibliography{references}

\appendix

\section{Experimental Result and Ablation Studies}\label{app:controls}

\subsection{Patch area and position}\label{app:budget}
We examine how patch area affects attack effectiveness and whether the attack depends on patch position.
We conduct the patch-area and patch-position ablations on FastWAM across all 40 LIBERO tasks, using 50 trials per task and keeping the remaining settings consistent with the standard experimental configuration.

\begin{figure}[!htb]
\centering
\includegraphics[width=.90\linewidth]{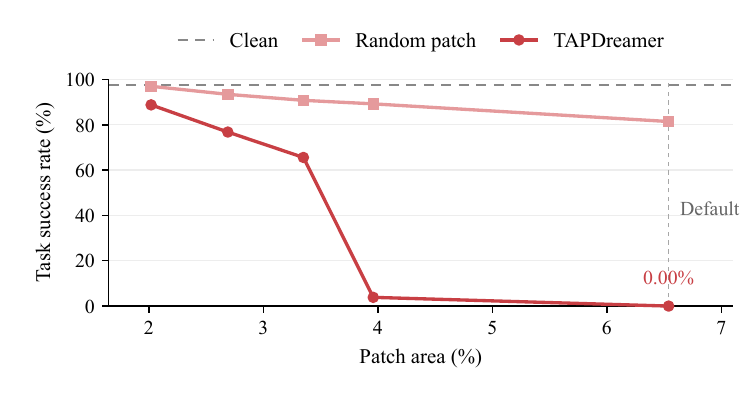}
\caption{Ablation study of patch area. Random patches use the same area and placement as \ourwork{}.}
\label{fig:area}
\end{figure}

As shown in Figure \ref{fig:area}, increasing patch area from 3.35\% to 3.96\% sharply reduces success from 65.60\% to 3.85\%, whereas the matched random patch changes success only from 90.80\% to 89.25\%. The sharp success-rate drop under \ourwork{}, compared with the small change under matched random patches, shows that the effect depends on optimized patch content rather than increased occlusion alone. 

We also study the effect of patch position. We consider four positions in the external camera view: upper left (0,0), upper right (0,143), lower left (143, 0), lower right (143, 143). Coordinates are $(y_0,x_0)$ in pixels. We consider two settings:  \textbf{reoptimized} patches are freshly optimized at the given position, and \textbf{moved} patches are copied from the patch optimized at the upper-left corner without update. 

Our results are presented in Table~\ref{tab:position}.
Reoptimizing a patch at each of the four tested positions achieves 0.00\% success, but moving the unchanged upper-left patch to the other three corners yields 93.50\%--95.00\% success, close to the corresponding random-patch results. Thus, effective patches can be optimized at different positions, but a patch optimized at one position does not transfer directly to another.

\begin{table}[!htb]
\centering\small
\setlength{\tabcolsep}{3pt}
\caption{Ablation study of patch position. Reoptimized patches are optimized at the given position; moved patches are copies of the unchanged upper-left patch. Success rates are percentages; subscripts report signed differences from Random at the same position, in percentage points.}
\label{tab:position}

\begin{tabular}{
    l
    r@{}l
    r@{}l
    r@{}l
    r@{}l
}
\toprule
\multirow{2}{*}{Setup}
& \multicolumn{2}{c}{Upper left}
& \multicolumn{2}{c}{Upper right}
& \multicolumn{2}{c}{Lower left}
& \multicolumn{2}{c}{Lower right} \\
& \multicolumn{2}{c}{$(0,0)$}
& \multicolumn{2}{c}{$(0,143)$}
& \multicolumn{2}{c}{$(143,0)$}
& \multicolumn{2}{c}{$(143,143)$} \\
\midrule

Random
& 81.45 & 
& 97.55 & 
& 96.20 & 
& 93.80 &  \\

Reoptimized
& 0.00 & \drop{81.45}
& 0.00 & \drop{97.55}
& 0.00 & \drop{96.20}
& 0.00 & \drop{93.80} \\

Moved
& 0.00  & \drop{81.45}
& 93.50 & \drop{4.05}
& 94.85 & \drop{1.35}
& 95.00 & \gain{1.20} \\

\bottomrule
\end{tabular}%
\end{table}

\makeatletter
\setlength{\@fptop}{0pt}
\setlength{\@fpsep}{12pt}
\setlength{\@fpbot}{0pt plus 1fil}
\makeatother

\subsection{Direct transfer of VLA patches}\label{app:vla_transfer}

To further broaden the attack spectrum beyond our design, we explore the potential of existing VLA attacks. Due to the incompatibility of the VLA and WAM paradigms, we run only the transfer setting, i.e., the attack targets a VLA victim and transfers the optimized patches to a WAM victim.

Table~\ref{tab:vla_transfer} reports direct transfer to FastWAM of patches constructed by UADA~\citep{r8} and UPA-RFAS~\citep{r4}.

We train UPA-RFAS and UADA patches on their respective source VLAs using the original attack settings, with the patch area matched to \ourwork{}. We then transfer these patches directly to FastWAM without further optimization. Overall success is 76.55\% for UPA-RFAS and 75.85\% for UADA, compared with 81.45\% for the random patch: reductions of 4.90 and 5.60 percentage points, respectively. \ourwork{} reduces success to 0\% across all four suites.

\begin{table}[!htbp]
\centering\small
\setlength{\tabcolsep}{3pt}
\caption{Direct transfer of VLA patches to FastWAM. Values are LIBERO task success rates (percent); subscripts report signed differences from Clean in the same suite or Overall column, in percentage points. Clean, Random, and \ourwork{} use the reference results in Table~\ref{tab:main}.}
\label{tab:vla_transfer}

\begin{tabular}{
    l
    r@{}l
    r@{}l
    r@{}l
    r@{}l
    r@{}l
}
\toprule
\multirow{2}{*}{Setup}
& \multicolumn{10}{c}{LIBERO} \\
\cmidrule(lr){2-11}
& \multicolumn{2}{c}{Spatial}
& \multicolumn{2}{c}{Object}
& \multicolumn{2}{c}{Goal}
& \multicolumn{2}{c}{Long-hor.}
& \multicolumn{2}{c}{Overall} \\
\midrule

Clean
& 98.00 & 
& 99.60 & 
& 97.60 & 
& 95.40 & 
& 97.65 &  \\

Random
& 85.60 & \drop{12.40}
& 78.00 & \drop{21.60}
& 90.00 & \drop{7.60}
& 72.20 & \drop{23.20}
& 81.45 & \drop{16.20} \\

UPA-RFAS
& 84.20 & \drop{13.80}
& 76.80 & \drop{22.80}
& 77.40 & \drop{20.20}
& 67.80 & \drop{27.60}
& 76.55 & \drop{21.10} \\

UADA
& 82.00 & \drop{16.00}
& 77.00 & \drop{22.60}
& 77.20 & \drop{20.40}
& 67.20 & \drop{28.20}
& 75.85 & \drop{21.80} \\

\ourwork{}
& 0.00 & \drop{98.00}
& 0.00 & \drop{99.60}
& 0.00 & \drop{97.60}
& 0.00 & \drop{95.40}
& 0.00 & \drop{97.65} \\

\bottomrule
\end{tabular}%
\end{table}

\section{How the Latent Shift Changes Action}\label{app:latent_action}
\subsection{Attention-restoration interventions}\label{app:restoration}
We want to understand which patch-induced changes within attention contribute to perturbation propagation, particularly the roles of attention weights $A$ and value vectors $V$ in downstream representation and action changes. To investigate this, we conduct three restoration interventions: restoring the entire attention branch, restoring only the attention weights $A$, and restoring only the value vectors $V$. Each intervention retains the patched residual branch and replaces only the specified component.

We measure the remaining change relative to the clean output, normalized by the change without restoration. For latent retention, both changes are measured using the L2 norm over the entire latent, including the patch footprint; action retention uses the corresponding L2 ratio for action outputs. The random patch is evaluated with the same latent-retention metric. Because attention reconstruction uses a different numerical precision from model inference, the patched reference without restoration is computed through the same reconstruction pathway. These ratios measure retained changes, not task success rates.

Table~\ref{tab:restore} shows that all three interventions leave approximately 33\% of the latent change and 32\% of the action change induced by \ourwork{}. In comparison, random-patch latent retention remains approximately 1.000. Thus, restoring the attention branch, its weights, or its values removes most of the \ourwork{}-induced latent and action changes, while having little effect on the random patch.

\begin{table}[!htbp]
\centering\small
\setlength{\tabcolsep}{5pt}
\caption{Fraction of patch-induced change retained after attention restoration. Latent and action columns report \ourwork{}; the last column reports random-patch latent retention. Values are medians over 48 states.}
\label{tab:restore}
\begin{tabular}{lrrr}
\toprule
Restored component & \shortstack{Latent\\retained} & \shortstack{Action\\retained} & \shortstack{Random latent\\retained} \\
\midrule
Restore clean attention branch & 0.3327 & 0.321 & 1.000 \\
Weights A restored to clean & 0.3364 & 0.320 & 1.000 \\
Values V restored to clean & 0.3327 & 0.321 & 1.000 \\
\bottomrule
\end{tabular}
\end{table}

\FloatBarrier
\subsection{Optimization trajectories}\label{app:trajectory}
To examine when shared perturbation structure emerges, we track optimization across three initializations on 48 observations. Within 25 steps, template agreement reaches at least 0.964 and effective rank falls below 1.39, while source displacement is only about 12\% of its final magnitude. Subsequent optimization mainly amplifies displacement on both source and held-out observations while maintaining high alignment.

\begin{figure}[!htb]
\centering
{\footnotesize\textcolor[HTML]{4477AA}{\rule[.5ex]{1.3em}{1pt}}~Initialization A\qquad\textcolor[HTML]{EE9944}{\rule[.5ex]{1.3em}{1pt}}~Initialization B\qquad\textcolor[HTML]{228833}{\rule[.5ex]{1.3em}{1pt}}~Initialization C}\par\vspace{2pt}
\includegraphics[width=.82727\linewidth]{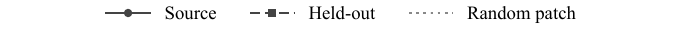}\par\vspace{2pt}
\begin{minipage}[t]{.49\linewidth}
\centering\vspace{0pt}
\includegraphics[width=\linewidth]{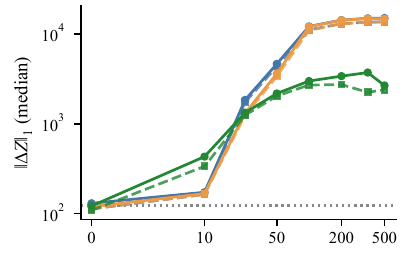}\par\vspace{1pt}
{\centering\small (a) Latent L1 displacement\par}
\end{minipage}\hfill
\begin{minipage}[t]{.49\linewidth}
\centering\vspace{0pt}
\includegraphics[width=\linewidth]{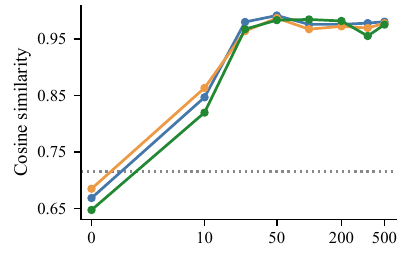}\par\vspace{1pt}
{\centering\small (b) Template agreement\par}
\end{minipage}
\par\vspace{4pt}
\begin{minipage}[t]{.49\linewidth}
\centering\vspace{0pt}
\includegraphics[width=\linewidth]{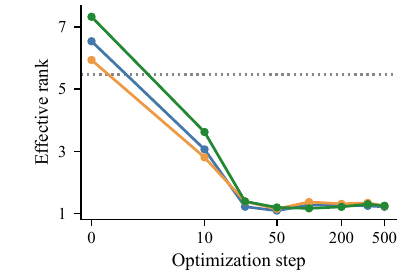}\par\vspace{1pt}
{\centering\small (c) Effective rank\par}
\end{minipage}\hfill
\begin{minipage}[t]{.49\linewidth}
\centering\vspace{0pt}
\includegraphics[width=\linewidth]{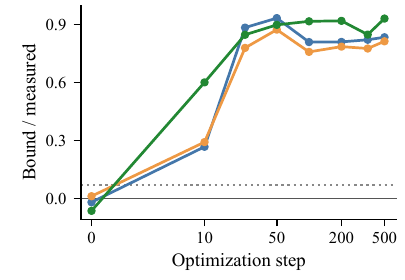}\par\vspace{1pt}
{\centering\small (d) Frozen-template L1 bound\par}
\end{minipage}
\caption{Evolution of perturbation structure across observations during optimization. For three initializations, the panels show displacement on source and held-out observations, source-template agreement, the effective rank of 48 perturbation fields, and the ratio of the frozen source-template lower bound to measured L1 displacement. In (a), solid and dashed curves denote source and held-out states, respectively. Dotted lines show results with random patches.}
\label{fig:trajectory_full}
\end{figure}

Figure~\ref{fig:trajectory_full}(a) shows that displacement grows on both source and held-out observations, indicating that the amplified perturbation extends beyond the source states. The early rise in template agreement in (b) and the effective rank approaching one in (c) indicate that a common direction emerges early and becomes dominant across observations. The high lower-bound ratio in (d) shows that the source-derived template accounts for a substantial fraction of held-out displacement even after subtracting the residual. Together, these trajectories indicate that optimization first establishes a shared direction and then mainly increases its magnitude.

\paragraph{Metrics and legend.}
At each optimization checkpoint in Figure~\ref{fig:trajectory_full}, we measure the final-latent change $\Delta Z_i$ on the 356 spatial cells outside the projected patch footprint. The source template $\overline{\Delta Z}_{\mathrm{src}}$ is the mean displacement field over 12 source states at that checkpoint and is held fixed when evaluating the 36 held-out states. Panel (a) reports the median $\|\Delta Z_i\|_1$ separately for the source and held-out states. Panel (b) reports the median cosine similarity $\langle\overline{\Delta Z}_{\mathrm{src}},\Delta Z_i\rangle/(\|\overline{\Delta Z}_{\mathrm{src}}\|_2\|\Delta Z_i\|_2)$ over held-out states, with displacement fields flattened into vectors.

For panel (c), we stack the 48 flattened displacement fields as rows of a matrix. Its effective rank is $\exp(-\sum_j q_j\log q_j)$, where $q_j=\sigma_j^2/\sum_k\sigma_k^2$ and $\sigma_j$ are its singular values (with $0\log0=0$). A value near one indicates that the displacement fields largely share one direction. Panel (d) reports the median ratio $(\|\overline{\Delta Z}_{\mathrm{src}}\|_1-\|\Delta Z_i-\overline{\Delta Z}_{\mathrm{src}}\|_1)/\|\Delta Z_i\|_1$ over held-out states: the frozen-template lower bound in Equation~\eqref{eq:templatebound} divided by the measured L1 displacement.

Colors identify the three initializations in all panels. Only panel (a) uses solid lines with circles for source states and dashed lines with squares for held-out states; panels (b)--(d) show one summary curve per initialization. Gray dotted horizontal lines show the corresponding random-patch results, rather than an additional initialization.

\subsection{Behavioral effects}\label{app:behavior}
The preceding analyses show how patch-induced perturbations propagate through attention and how optimization establishes a shared direction before amplifying its magnitude. We now ask how these representation changes affect action commands and, in turn, lead to task failure.

\begin{figure}[!htb]
\centering
\begin{minipage}{.85\linewidth}
\centering
\begin{minipage}[b]{.665\linewidth}
\centering
\includegraphics[width=\linewidth]{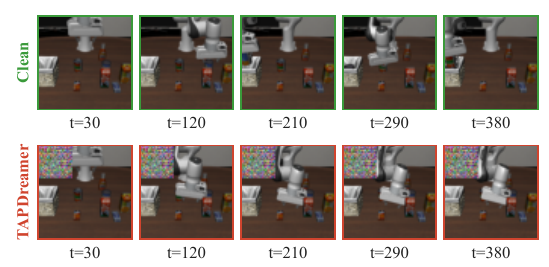}\par\vspace{2pt}
\parbox[t][2.2\baselineskip][t]{\linewidth}{\raggedright\small (a) Put the alphabet soup and tomato sauce in the basket\par}
\end{minipage}\hfill
\begin{minipage}[b]{.325\linewidth}
\centering
\includegraphics[width=\linewidth]{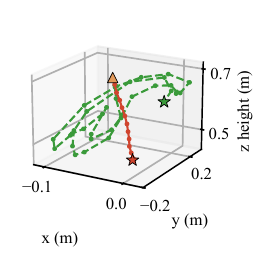}\par\vspace{2pt}
\parbox[t][2.2\baselineskip][t]{\linewidth}{\raggedright\small (b) End-effector trajectory\par}
\end{minipage}
\end{minipage}
\caption{Example of closed-loop behavior under \ourwork{}. Left: observation sequences under clean and perturbed conditions. Right: corresponding three-dimensional end-effector position trajectories. Green dashed and red solid paths denote Clean and \ourwork{}, respectively. The triangle marks the starting position and stars mark the endpoints; $t$ counts environment execution steps. Positions are in meters in the world coordinate frame.}
\label{fig:execution}
\end{figure}

\paragraph{Visible execution failures.}
The same representation shift can affect different actions as a task unfolds. In Figure~\ref{fig:execution}, the patch drives the end effector below its clean trajectory and prevents task completion. We use same-state action comparisons, internal restoration, and patch removal to connect the shift to commands and physical consequences.

\paragraph{Immediate action changes.}
Holding the physical state fixed, we compare actions from clean and patched observations before their trajectories diverge. In FastWAM, the patch suppresses lifting, biases pose commands downward, and shifts gripper commands toward opening. A fit in which each patched six-dimensional pose command mixes its clean counterpart with a shared target explains 78.2\% of the variation across 1,500 paired commands. These are state-level measurements, not independent episodes. The same bias can prevent a lift, lose an object after grasping, or move the end effector away from its intended position, depending on the task stage. These action patterns emerge from the encoder-level objective, which specifies no target action.

\paragraph{Action pathways across models.}
The disturbed visual representation also reaches actions through different downstream designs. In Motus, restoring the branch that represents visual world dynamics recovers about 93\%--99\% of the action change, whereas restoring the visual-language understanding branch recovers about 10\%--27\%. In DreamWAM-joint, restoring predicted future representations also substantially recovers actions. These interventions link encoder interference to control while showing that its route depends on the model's action pathway.

\paragraph{Continued exposure and recovery.}
Gripper commands shift toward closing at the first clean replan after removal, indicating that the immediate command bias depends on continued visual interference. Whether the task recovers then depends on the physical state: some pose deviations can be corrected, while dropped or misplaced objects and missed manipulation opportunities may persist. In the duration sweep, some tasks finish after removal following 20 corrupted replans, whereas continuous exposure maintains strong failure. Repeated exposure lets the fixed patch keep injecting the same latent change as execution accumulates physical consequences.

\end{document}

%% file: sec/intro.tex
\section{Introduction}
\label{sec:intro}
\begin{figure}[t]
\centering
\begingroup
\setlength{\unitlength}{\dimexpr\linewidth/1470\relax}
\begin{picture}(1470,651.005)
\put(0,0){\includegraphics[width=393\unitlength,height=651.005\unitlength]{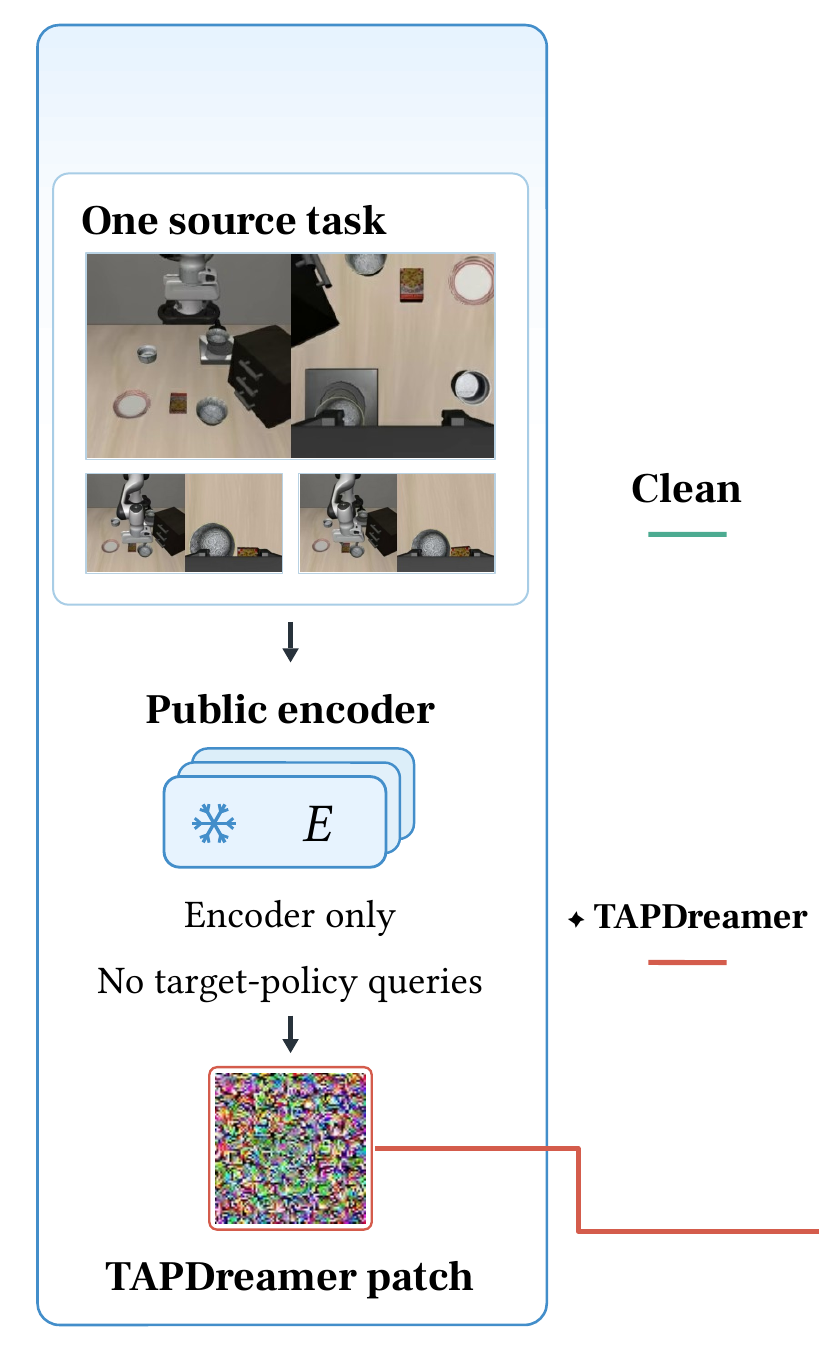}}
\put(393,0){\includegraphics[width=1076.991\unitlength,height=651.005\unitlength]{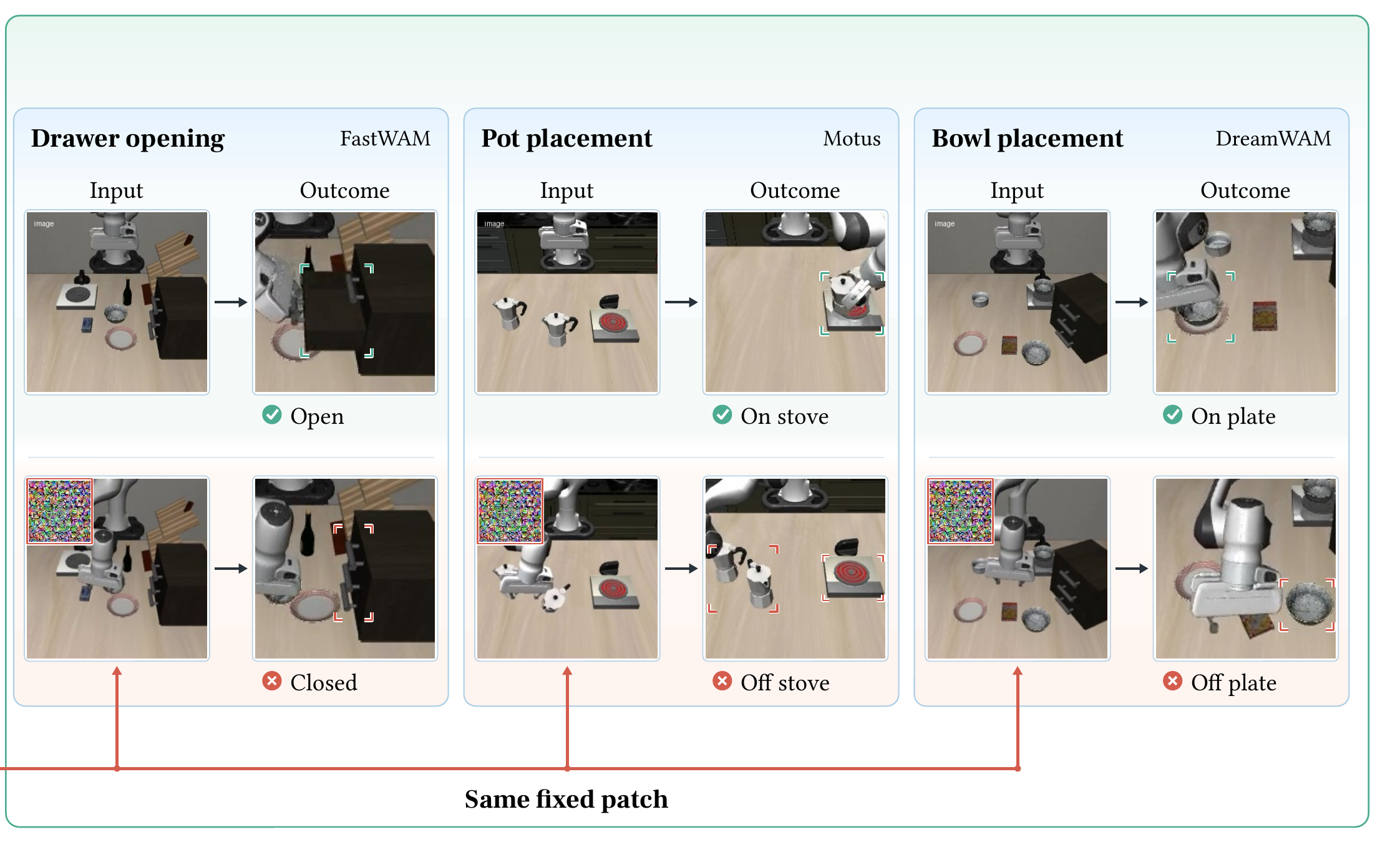}}
\put(25,602){\makebox(0,0)[bl]{\fontsize{7.5}{9}\selectfont\textbf{(a) Optimize once}}}
\put(408,602){\makebox(0,0)[bl]{\fontsize{7.5}{9}\selectfont\textbf{(b) Reuse the same patch}}}
\end{picture}
\endgroup
\caption{\ourwork{} constructs a patch using a public visual encoder and observations from one source task, then freezes it for deployment across tasks and WAMs within each benchmark. Examples compare clean and patched execution on drawer opening with FastWAM, pot placement with Motus, and bowl placement with DreamWAM.}
\label{fig:teaser}
\end{figure}

\begin{figure}[t]
\centering
\includegraphics[width=\linewidth]{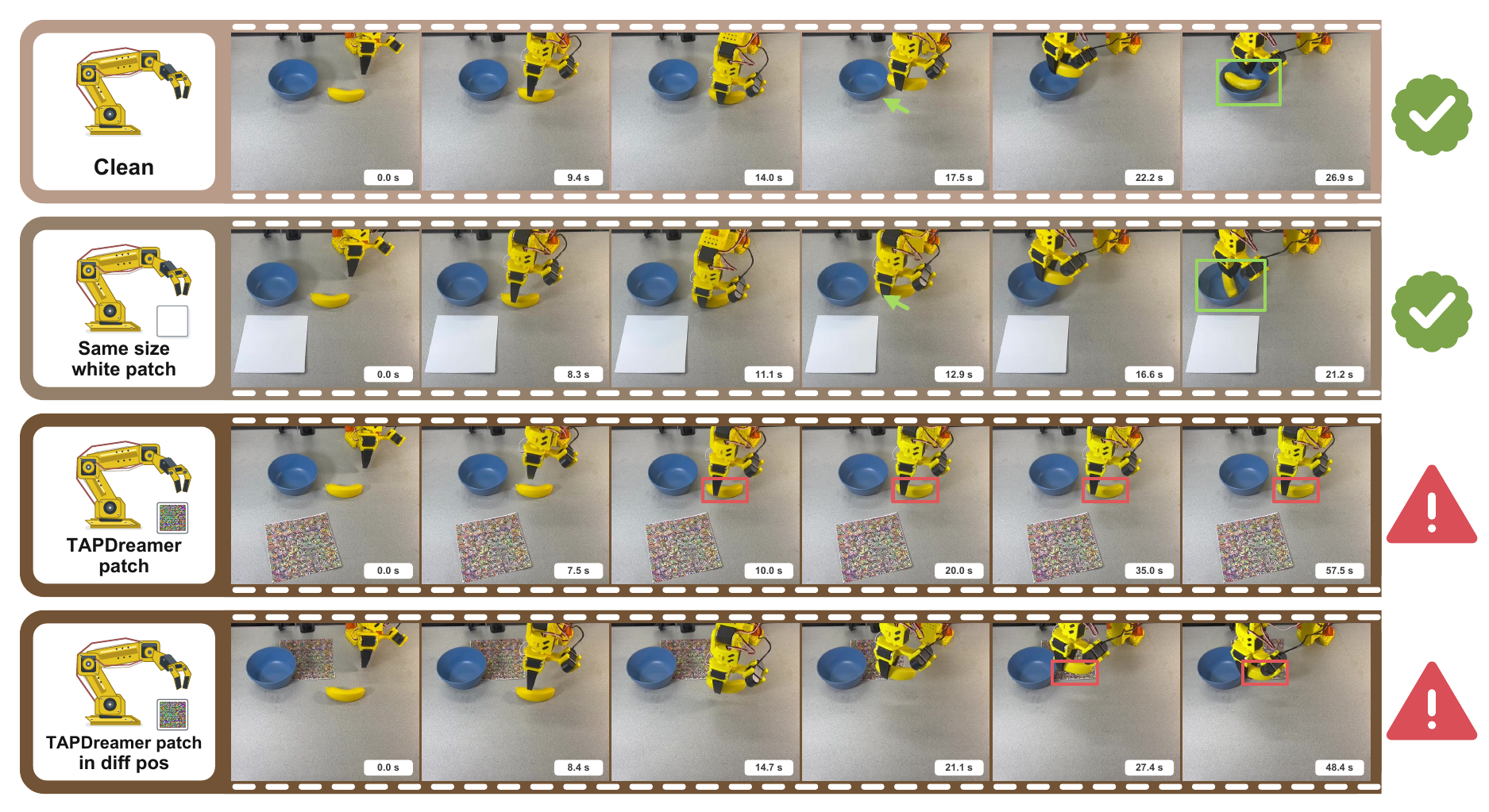}
\caption{Real-world FastWAM executions on SO-101 for ``Pick up the banana and place it in the bowl.'' Rows show clean execution, a same-size white patch, and TAPDreamer at two positions.}
\label{fig:realworld_sequences}
\end{figure}

World Action Models (WAMs) are emerging as a powerful foundation for general-purpose robotic control. Rather than predicting actions from the current observation alone, a WAM learns how the visual world evolves together with the robot's actions. This world-modeling objective captures objects, spatial relationships, and motion, and can support action generation either through explicit future imagination~\citep{ref_unipi,ref_robodreamer} or through representations learned during video prediction~\citep{r1}. By unifying visual prediction and control, WAMs promise policies that can reason about the consequences of their actions and replan as the environment changes~\citep{ref_uwm,ref_dreamzero}.

This coupling, however, also creates a security risk. Current WAM pipelines implicitly trust the visual representation on which world prediction and action generation are built. This assumption is fragile. Cameras expose a direct, externally accessible input channel. Moreover, visual encoders are often reused across tasks and models, making the exposed channel transferable among tasks. A perturbation that corrupts this shared interface can therefore bypass model-specific action heads. If the channel remains exposed, a perturbation can be injected consistently. In this paper, we investigate whether an attacker can exploit this vulnerability and launch attacks against WAMs that transfer to unseen tasks and models.

We propose \ourwork{} (illustrated in Figure \ref{fig:teaser}), a \emph{transferable attack} that applies a patch on visual inputs to WAMs.
\ourwork{} maximizes the displacement of the final latent produced by visual encoder. The search uses a few demonstration frames from a single source task. Once the patch is found, the patch pixels, area, and position are frozen and the same artifact is deployed unchanged throughout each episode and across target tasks and models. Throughout the attack, \ourwork{} does not require action labels, task rewards, simulator, victim-model weights, or policy outputs, making \ourwork{} simple and inexpensive to initiate.

We evaluate \ourwork{} on 40 LIBERO tasks and all 50 RoboTwin tasks. \ourwork{} consistently outperforms attack baselines. On FastWAM, a single frozen patch covering approximately $6.5\%$ of the input reduces task success from $97.65\%$ to $0.00\%$ on LIBERO and from $90.86\%$ to $0.00\%$ on RoboTwin. Random patches retain $81.45\%$ and $79.20\%$ success, respectively. The same frozen patches also transfer across WAM architectures without retraining or target-model queries: success falls from $84.55\%$ and $86.10\%$ with matched random patches to $1.45\%$ and $1.00\%$ on DreamWAM-uncond and DreamWAM-joint, and from $82.40\%$ to $10.60\%$ on Motus.
We further validate \ourwork{} in real-world deployment on an SO-101 robot, with representative FastWAM executions shown in Figure~\ref{fig:realworld_sequences}.
We also trace how a fixed patch causes failure during closed-loop control. 
Our results show that changing a small set of pixels distorts visual features across the scene, including those used to select actions, and thus leads to reduced task success rate.

%% file: sec/setup.tex
\section{World Action Model, Threat Model, and \ourwork{}}
\label{sec:setup}

\subsection{World Action Model}
Let $s_t$ denote the robot and environment state at step $t$, $o_t$ the camera observations, $q_t$ the robot's proprioceptive state, $\ell$ the task instruction, and $\mathcal H_t$ the available history. For a world model $m$, image preprocessing $G_m$ produces the visual input $x_t=G_m(o_t)$, and the visual encoder produces $Z_t=E(x_t)$. The downstream model constructs a context $C_t=F_\theta(Z_t,\ell,q_t,\mathcal H_t),$
where $\theta$ denotes the learnable parameters of the corresponding module. 

For a WAM that predicts future visual latents before generating actions, we denote visual predictions and generated action chunks as
\begin{equation}
\hat Z_{t+1:t+K_v}\sim W_\theta(\,\cdot\mid C_t\,),\qquad
a_{t:t+K_a-1}\sim A_\theta(\,\cdot\mid C_t,\hat Z_{t+1:t+K_v}\,),
\label{eq:wam}
\end{equation}
where $K_v$ and $K_a$ are the visual prediction and action horizons. Here, $W_\theta$ and $A_\theta$ denote conditional distributions over future latents and action chunks.
During closed-loop execution, the robot executes the first $K_e$ actions of each chunk and receives new observations:
\begin{equation}
s_{t+K_e}=\mathcal F(s_t,a_{t:t+K_e-1}),\qquad o_{t+K_e}=O(s_{t+K_e}),
\label{eq:loop}
\end{equation}
where $1\leq K_e\leq K_a$, $\mathcal F$ is the environment transition, and $O$ is the observation function. The WAM then encodes the new observations and generates the next chunk. A visual perturbation can therefore affect both the current action and subsequent observations through execution-induced state changes.

\subsection{Threat Model}
\paragraph{Attacker's objective.}
We consider an attacker who aims to reduce the task success rate of a WAM-controlled robot by modifying a small region of its visual input. This is an untargeted, test-time attack.

\paragraph{Attacker's knowledge and access.}
The attacker has white-box access to a publicly released visual encoder, including its weights and gradients, and a small set of public demonstration images from one source task. We assume that the attacker knows the relevant visual preprocessing, including image normalization and camera-view composition. Patch optimization requires only the WAM's encoder and a small set of demonstration frames from a single source task. The attacker does not require access to the downstream world model or action decoder or their gradients, action labels, or task rewards for pixel optimization.

\paragraph{Attacker's capabilities.}
The attacker can replace pixels within a fixed local region of the visual input before encoding. Let $M$ be a fixed binary mask and $p$ the patch. The attacked observation is
\begin{equation}
T_p(x_t)=(1-M)\odot x_t+M\odot p,
\label{eq:patch}
\end{equation}
where $\odot$ represents elementwise multiplication. The attacker chooses the patch content. At deployment, the patch content, position, and area remain unchanged. The same patch is reused across target tasks and models within each benchmark. The attack does not manipulate pixels outside the mask, language instructions, proprioceptive inputs, or policy parameters. The attacker cannot directly set the robot's actions or the environment state.

\begin{figure}[t]
\centering
\begingroup
\setlength{\unitlength}{\dimexpr\linewidth/1470\relax}
\begin{picture}(1470,480)
\put(0,0){\includegraphics[width=937\unitlength,height=480\unitlength]{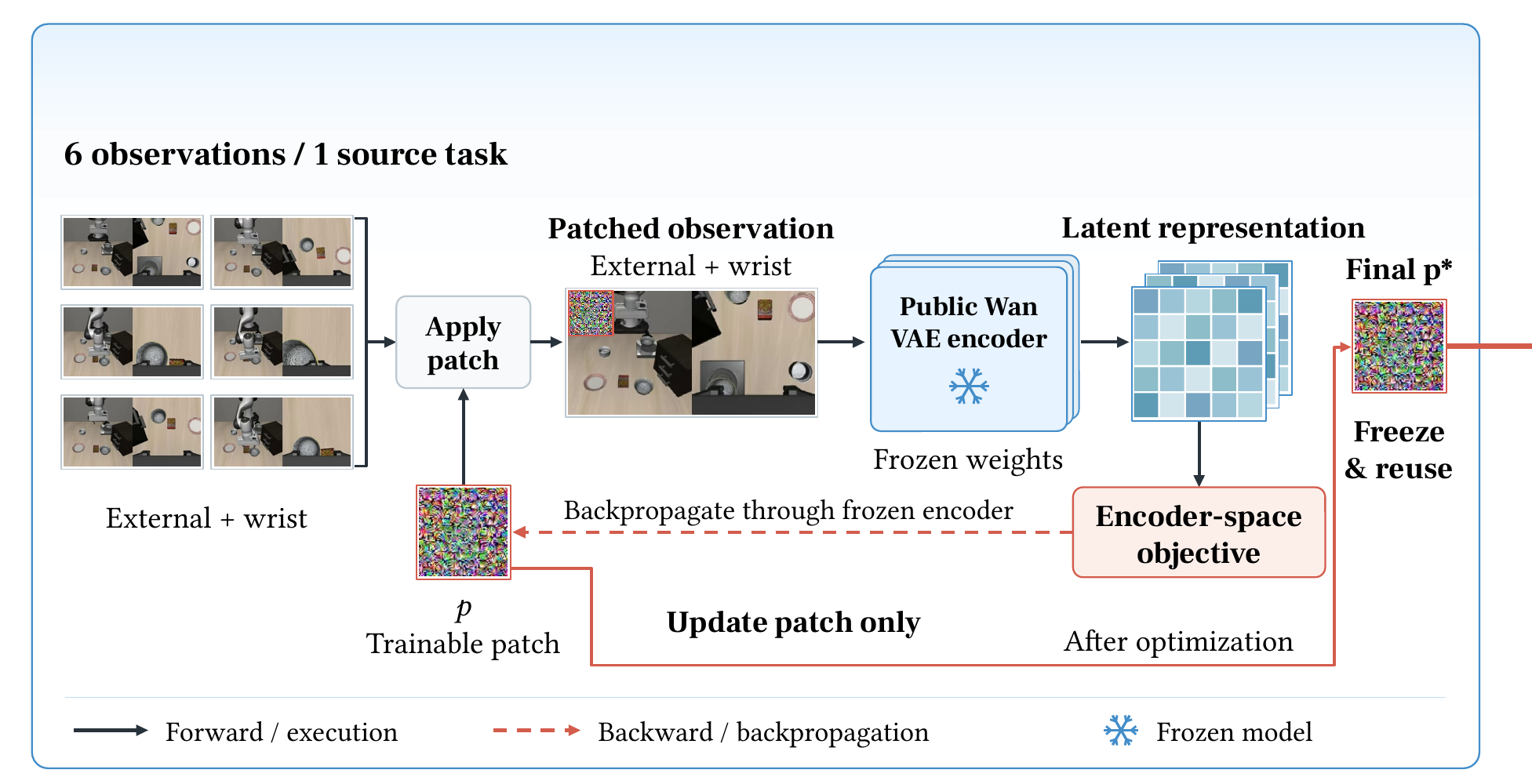}}
\put(937,0){\includegraphics[width=532.991\unitlength,height=480\unitlength]{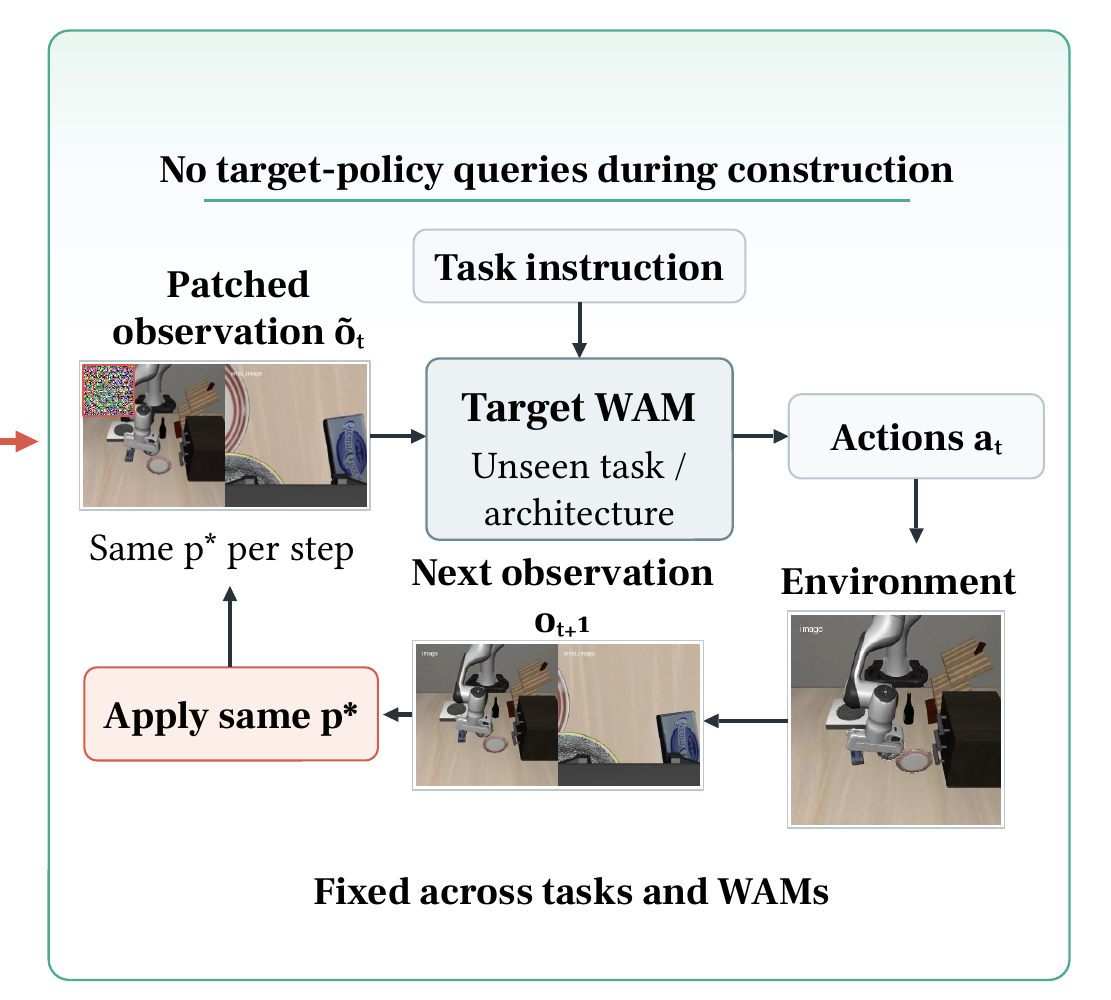}}
\put(33,430.134){\makebox(0,0)[bl]{\small\textbf{(a) Offline patch construction}}}
\put(974,430.134){\makebox(0,0)[bl]{\small\textbf{(b) Fixed-patch deployment}}}
\end{picture}
\endgroup
\caption{Overview of \ourwork{}. (a) Pixel updates maximize the latent displacement through a frozen public visual encoder. (b) The resulting patch is frozen and reused at each step across tasks and models within the same benchmark.}
\label{fig:overview}
\end{figure}

\begin{algorithm}[b]
\caption{TAPDreamer Patch Optimization}
\label{alg:tap}
\begin{algorithmic}[1]
\Require Frozen encoder $E$, source images $\{x_j\}_{j=1}^{J}$,
patch mask $M$, pixel range $[a,b]$, iterations $K$, learning rate $\eta$
\Ensure Optimized patch $p^\star$

\State $Z_j^0 \gets \Call{StopGradient}{E(x_j)},\ \forall j$
\State $p_0 \sim \mathcal{U}(a,b)$; initialize Adam state $\theta_0$
\For{$k \gets 0$ to $K-1$}
    \State $T_{p_k}(x_j) = (1-M)\odot x_j + M\odot p_k,\ \forall j$
    \Comment{Eq.~\eqref{eq:patch}}
    \State $\displaystyle
    \mathcal{L}_k \gets \frac{1}{J N_Z}
    \sum_{j=1}^{J}
    \left\|E(T_{p_k}(x_j))-Z_j^0\right\|_1$
    \State $(p_{k+1},\theta_{k+1})
    \gets \operatorname{Adam}
    (p_k,-\nabla_{p_k}\mathcal{L}_k,\theta_k;\eta)$
\EndFor
\State \Return $p_K$
\end{algorithmic}
\end{algorithm}

\subsection{Attack Design and Implementation}
\paragraph{Design rationale.}
The attack exploits the reuse of public visual encoders across WAMs with different downstream control architectures. The encoder output therefore provides an attack objective independent of a particular action interface. For local pixel changes to affect broader scene representations, they must exploit propagation pathways within the encoder. We maximize the difference between the encoder's clean and patched representations over the entire latent output, including all spatial positions and channels. The same patch is optimized over multiple source observations. 

\paragraph{Patch construction.}
\ourwork{} samples $J$ frames $\{x_j\}_{j=1}^{J}$ from different demonstrations of a single source task, caches their clean latents, and optimizes
\begin{equation}
\max_{p\in\mathcal P_{\mathrm{local}}} \frac{1}{JN_Z}\sum_{j=1}^{J}\norm{E(T_p(x_j))-E(x_j)}_1.
\label{eq:attack}
\end{equation}
Here, $N_Z$ is the total number of scalar elements in the final latent, and $\mathcal P_{\mathrm{local}}$ contains patches within the fixed mask that satisfy the valid pixel range. 
Algorithm~\ref{alg:tap} gives the construction procedure.
Figure~\ref{fig:overview} illustrates the overall pipeline.

%% file: sec/experiment.tex
\section{Experiments}\label{sec:experiments}
\subsection{Experimental Setup}\label{sec:experimental_setup}
\paragraph{Datasets and models.}
We conduct closed-loop evaluation on LIBERO and RoboTwin~\citep{r6,r7}. LIBERO contains four task categories (Spatial, Object, Goal, and Long-horizon) with 10 tasks each. RoboTwin contains 50 tasks. FastWAM is the default evaluation model. DreamWAM-uncond (DW-U), DreamWAM-joint (DW-J)~\citep{ref_dreamwam}, and Motus~\citep{ref_motus} are used for transfer evaluation within the same benchmark. Each model configuration retains its own action generation process and preprocessing.

\paragraph{Attack setup.} We run \ourwork{} using patches optimized on source tasks; cross-model transfer occurs within each benchmark. Patches are injected before encoding. On LIBERO, the external and wrist views form a $224\times448$ canvas; the main patch occupies an $81\times81$ region in the upper-left corner of the external view, covering 6.54\% of the input. The corresponding coverage on RoboTwin is 6.59\%. Patches remain visible throughout each trial in standard attack evaluations. 
Images used for optimization come from public LeRobot demonstrations. We filter episodes by source-task identifier, then select one frame from each of six episodes at temporal indices 5, 20, 40, 60, 15, and 35, respectively. When generating the patch, \ourwork{} sets $J=6$ with 500 Adam gradient-ascent updates on patch pixels, using a learning rate of 0.02. Patch pixels are initialized uniformly within $[-1,1]$. After each update, they are projected back to this range, with support enforced by the fixed mask. 

\paragraph{Real-world setup.}
We additionally evaluate FastWAM and DreamWAM on an SO-101 robot for ``Pick up the banana and place it in the bowl,'' using 50 trials per model and condition and retaining the original patch size.
The FastWAM patch is transferred to DreamWAM without modification or reoptimization.

\paragraph{Baselines.} We compare \ourwork{} with baselines to show its effectiveness. We evaluate the policy on unmodified observations to establish the \textbf{clean} baseline. For \textbf{random patch}, we replace the same image region with a random patch matched to the adversarial patch in position and area, to measure the effects of ordinary local occlusion and pixel replacement. 
Random patches are generated following a uniform distribution within admissible range of pixel values.
Extra ablation of transferred VLA attack is deferred to Appendix~\ref{app:vla_transfer}.

\begin{table}[t]
\centering
\caption{Closed-loop FastWAM success rates (percent) on complete task sets.
LIBERO and RoboTwin each use one fixed patch, optimized on
\texttt{Spatial t6} and \texttt{pick diverse bottles}, respectively. Small subscripts report the signed success-rate difference from Clean, in percentage points. LIBERO Overall averages the four suites.}
\label{tab:main}

\resizebox{\textwidth}{!}{%
\begin{tabular}{
    l
    r@{}l
    r@{}l
    r@{}l
    r@{}l
    r@{}l
    r@{}l
}
\toprule
\multirow{2}{*}{Setup}
& \multicolumn{10}{c}{LIBERO}
& \multicolumn{2}{c}{\multirow{2}{*}{RoboTwin}} \\
\cmidrule(lr){2-11}
& \multicolumn{2}{c}{Spatial}
& \multicolumn{2}{c}{Object}
& \multicolumn{2}{c}{Goal}
& \multicolumn{2}{c}{Long-hor.}
& \multicolumn{2}{c}{Overall}
& \multicolumn{2}{c}{} \\
\midrule

Clean
& 98.00 & \nodrop
& 99.60 & \nodrop
& 97.60 & \nodrop
& 95.40 & \nodrop
& 97.65 & \nodrop
& 90.86 & \nodrop \\

Random
& 85.60 & \drop{12.40}
& 78.00 & \drop{21.60}
& 90.00 & \drop{7.60}
& 72.20 & \drop{23.20}
& 81.45 & \drop{16.20}
& 79.20 & \drop{11.66} \\

\ourwork{}
& 0.00 & \drop{98.00}
& 0.00 & \drop{99.60}
& 0.00 & \drop{97.60}
& 0.00 & \drop{95.40}
& 0.00 & \drop{97.65}
& 0.00 & \drop{90.86} \\

\bottomrule
\end{tabular}%
}
\end{table}

\paragraph{Metrics.}
We evaluate attack effectiveness using closed-loop task \textbf{success rate} ($\mathrm{SR}$), defined as
\begin{equation}
\mathrm{SR}=\frac{N_{\mathrm{success}}}{N_{\mathrm{trials}}}\times100\%,
\label{eq:sr}
\end{equation}
where $N_{\mathrm{success}}$ is the number of successful trials and $N_{\mathrm{trials}}$ is the total number of trials. A trial is successful if the robot completes the task according to the benchmark's success criterion. 
Complete task sets contain both the source task and held-out tasks; cross-task transfer is evaluated on held-out tasks after freezing the patch.

\begin{table}[!b]
\centering\small
\setlength{\tabcolsep}{6pt}
\caption{Frozen-patch transfer across models within each benchmark. LIBERO results use the first optimization initialization. Success rates are percentages; small subscripts report signed differences from Clean for the same model and suite, in percentage points. DW-U and DW-J denote DreamWAM-uncond and DreamWAM-joint. LIBERO Overall averages the four suites.}
\label{tab:transfer}\label{tab:libero}

\begin{tabular}{
    ll
    r@{}l
    r@{}l
    r@{}l
}
\toprule
Victim & Suite
& \multicolumn{2}{c}{Clean}
& \multicolumn{2}{c}{Random}
& \multicolumn{2}{c}{\ourwork{}} \\
\midrule

\multicolumn{8}{l}{\textbf{LIBERO}} \\

DW-U & Spatial
& 96.20 & \nodrop
& 82.20 & \drop{14.00}
& 0.00  & \drop{96.20} \\

DW-U & Object
& 99.60 & \nodrop
& 96.00 & \drop{3.60}
& 0.00  & \drop{99.60} \\

DW-U & Goal
& 98.00 & \nodrop
& 90.00 & \drop{8.00}
& 5.80  & \drop{92.20} \\

DW-U & Long-horizon
& 96.20 & \nodrop
& 70.00 & \drop{26.20}
& 0.00  & \drop{96.20} \\

DW-U & Overall
& 97.50 & \nodrop
& 84.55 & \drop{12.95}
& 1.45  & \drop{96.05} \\

\addlinespace

DW-J & Spatial
& 99.40 & \nodrop
& 84.20 & \drop{15.20}
& 0.00  & \drop{99.40} \\

DW-J & Object
& 100.00 & \nodrop
& 99.20  & \drop{0.80}
& 0.00   & \drop{100.00} \\

DW-J & Goal
& 98.00 & \nodrop
& 88.00 & \drop{10.00}
& 4.00  & \drop{94.00} \\

DW-J & Long-horizon
& 93.40 & \nodrop
& 73.00 & \drop{20.40}
& 0.00  & \drop{93.40} \\

DW-J & Overall
& 97.70 & \nodrop
& 86.10 & \drop{11.60}
& 1.00  & \drop{96.70} \\

\midrule
\multicolumn{8}{l}{\textbf{RoboTwin}} \\

FastWAM & Overall
& 90.86 & \nodrop
& 79.20 & \drop{11.66}
& 0.00  & \drop{90.86} \\

Motus & Overall
& 86.62 & \nodrop
& 82.40 & \drop{4.22}
& 10.60 & \drop{76.02} \\

\bottomrule
\end{tabular}
\end{table}

\subsection{Experimental Results}
\paragraph{\ourwork{} reduces the success rate to zero on both benchmarks.}
We evaluate FastWAM on all four LIBERO suites with 10 tasks each ($4\times10$ tasks), using 50 trials per task, and on all 50 RoboTwin tasks, using 100 trials per task. On LIBERO, FastWAM achieves 97.65\% success with clean inputs, whereas the frozen \ourwork{} patch reduces performance to 0.00\%. The same pattern holds on RoboTwin, where success decreases from 90.86\% to 0.00\%. Table~\ref{tab:main} shows that success is 0.00\% in all four LIBERO suites, covering different task configurations.

\paragraph{\ourwork{} causes a much larger performance degradation than a matched random patch.}
On LIBERO, the random patch retains an 81.45\% success rate, compared with 0.00\% for \ourwork{}. On RoboTwin, the corresponding success rates are 79.20\% and 0.00\%, respectively. 
Thus, the performance degradation caused by \ourwork{} cannot be attributed solely to obscuring the same image region; it depends on the optimized content of the \ourwork{} patch.

\begin{figure}[!tbp]
\centering
\includegraphics[width=\linewidth]{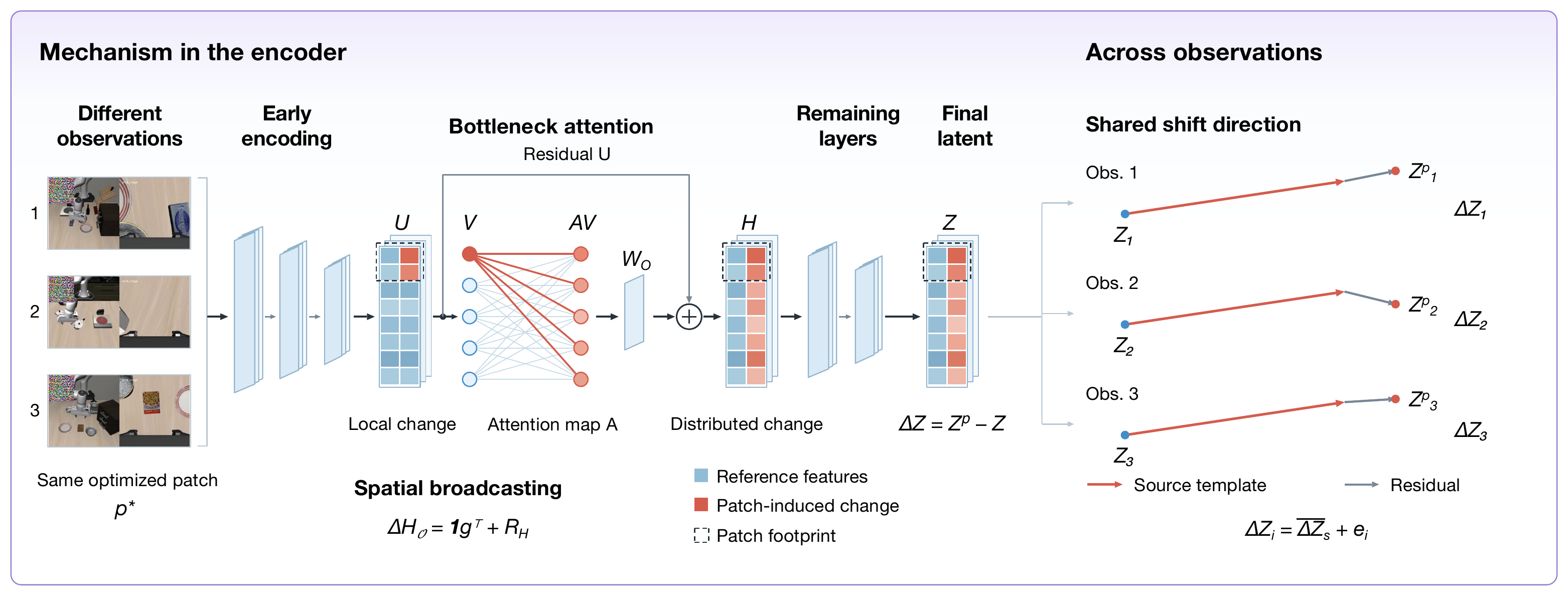}
\caption{Mechanism overview of \ourwork{}. Alocal patch-induced change spreads through bottleneck attention and propagates to the final latent. The changes across observations share a source-template component, with observation-specific residuals.}
\label{fig:mechanism_overview}
\end{figure}

\paragraph{\ourwork{} transfers across downstream control architectures.}
Keeping the patch fixed and changing the victim model reduces average success to 1.45\% on DreamWAM-uncond, 1.00\% on DreamWAM-joint, and 10.60\% on Motus. The corresponding matched random patches retain 84.55\%, 86.10\%, and 82.40\% success. Table~\ref{tab:transfer} reports results by benchmark, model, and suite. Transfer uses no target-model feedback for reoptimization.

DreamWAM-uncond does not read explicitly generated future frames to predict actions, yet it is still strongly affected. In these models, explicit test-time future-video generation is therefore not necessary for the attack.

\paragraph{Real-world results.}
\ourwork{} reduces FastWAM's success rate from $92\%$ to $0\%$ and DreamWAM's from $80\%$ to $20\%$; representative FastWAM executions are shown in Figure~\ref{fig:realworld_sequences}.

We defer more experimental results along with ablation studies about patch variations to Appendix~\ref{app:controls}.

\begin{table}[t]
    \centering
    \caption{Fraction of squared representation change outside the projected patch footprint: $\|\Delta X_O\|_F^2/\|\Delta X\|_F^2$, where $\Delta X=X^p-X$, $X\in\{U,H,Z\}$ denotes attention input, attention output, and final latent, respectively, and $O$ denotes positions outside the footprint at the corresponding layer. Values are medians over 48 states.}
    \label{tab:energy}
    \begin{tabular}{l c c c }
        \toprule
        & Attention input & Attention output & Final latent \\
        \midrule
        Random     & 0.007 & 0.008 & 0.013 \\
        \ourwork{} & 0.007 & 0.708 & 0.794 \\
        \bottomrule
    \end{tabular}
\end{table}

%% file: sec/mechanism.tex
\section{Mechanism and Behavioral Analysis}\label{sec:mechanism}

This section investigates how a local pixel change by \ourwork{} results in deviations in action execution. 
Figure~\ref{fig:mechanism_overview} summarizes impacts on latents due to \ourwork{}.

\subsection{A local patch becomes a distributed latent shift}\label{sec:propagation}
\paragraph{Attention spreads the change introduced by \ourwork{}.}
We compare clean and patched activations throughout the VAE encoder. Its bottleneck contains spatial self-attention on a $14\times28$ grid. We call the projection of the patch region onto a feature grid the \emph{patch footprint}. Table~\ref{tab:energy} reports the fraction of squared magnitude of representation change outside this footprint. 
For \ourwork{}, it rises from less than 1\% at the attention input to about 71\% at its output and 79\% in the final latent. A random patch of the same size leaves changes largely within the footprint. The attention layer helps the optimized local change reach distant positions.

\begin{figure}[t]
\rmfamily
\renewcommand{\ourwork}{TAPDreamer}
\centering
\begingroup
\setlength{\tabcolsep}{0pt}
\begin{tabular}{@{}c@{\hspace{.02in}}c@{\hspace{.02in}}c@{\hspace{.03in}}c@{}}
\raisebox{.20in}{\includegraphics[width=.38in]{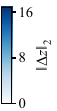}} &
\raisebox{.2446694in}{\includegraphics[width=1.29in]{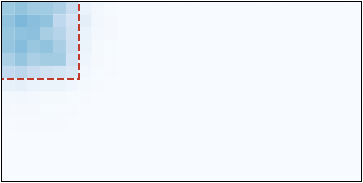}} &
\raisebox{.2446694in}{\includegraphics[width=1.29in]{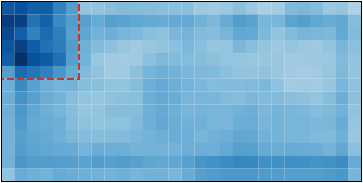}} &
\includegraphics[width=2.47in]{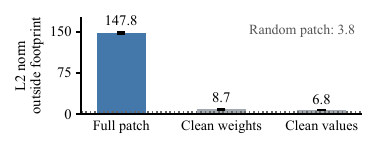} \\[2pt]
& {\small (a) Random patch}
& {\small (b) \ourwork{} patch}
& {\small (c) Attention restoration}
\end{tabular}
\par\endgroup
\caption{Spatial spread and attention restoration. (a,b) Per-position final-latent L2 changes with a shared color scale; dashed boxes mark the patch footprint. (c) Mean L2 change outside the footprint over 48 states, with 95\% confidence intervals. The dotted line denotes the random patch.}
\label{fig:spatial}
\end{figure}

\paragraph{Attention weights and values carry the change.}
To identify what carries this change, write the clean attention output after residual addition as $H=U+AVW_O$, where $U$ is the residual input, $A$ the attention weights, $V$ the value vectors, and $W_O$ the output projection. Rows index spatial positions; for multiple heads, $AV$ denotes their concatenated outputs. Let $X^p$ denote the patched quantity and $\Delta X=X^p-X$, where $X\in\{H, U, V, A\}$. We then have
\begin{equation}
\Delta H=\Delta U+(A\Delta V+\Delta A V+\Delta A\Delta V)W_O.
\label{eq:decomposition}
\end{equation}
The terms capture changed values, changed attention weights, and their joint change, respectively. 
We observe that the joint change term $\Delta A\Delta V$ is the largest measured component for \ourwork{} but is small for a random patch. 
This points to coordinated changes in weights and values, rather than greater attention to the patch alone.

\paragraph{Restoration tests the attention pathway.}
In Figure~\ref{fig:spatial}(c), Full patch denotes the patched pass without intervention; Clean weights and Clean values restore $A$ and $V$, respectively, to their clean counterparts. Restoring the attention branch, weights, or values to their clean counterparts while retaining the patched residual branch substantially reduces the latent and action changes. Perturbations outside the patch footprint are strongly suppressed, while the remaining perturbations are concentrated near the patch. The same interventions have little effect on the random patch. Figure~\ref{fig:spatial} shows the spatial changes and their magnitude outside the footprint; Table~\ref{tab:restore} in Appendix~\ref{app:restoration} reports the retained effects after each intervention.

\par\begin{table}[t]
\centering\small
\setlength{\tabcolsep}{5pt}
\caption{Consistency of final-latent changes outside the patch footprint. CV denotes coefficient of variation.}
\label{tab:stability}
\begin{tabular}{lrr}
\toprule
Metric & \ourwork{} & Random \\
\midrule
Cross-task cosine similarity & 0.971 & 0.603 \\
Magnitude CV & 0.056 & 0.071 \\
Template cosine similarity & 0.981 & 0.716 \\
Relative L1 residual & 0.206 & 0.809 \\
L1 bound ratio & 0.833 & 0.069 \\
\bottomrule
\end{tabular}
\end{table}\par

\begin{figure}[t]
\centering
\includegraphics[width=.85\linewidth]{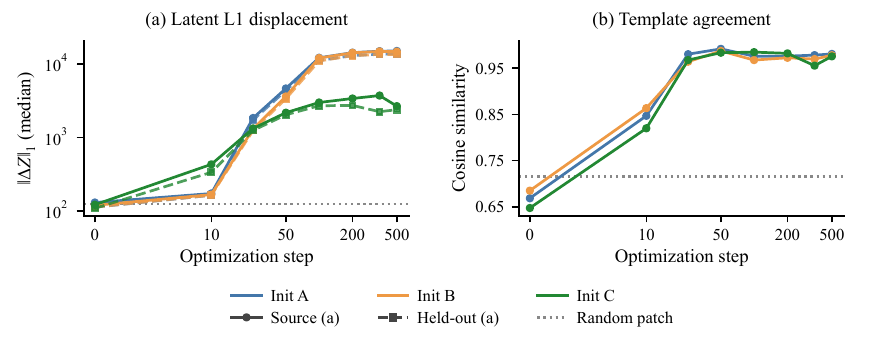}
\caption{Optimization trajectories for three initializations: (a) latent L1 displacement and (b) source-template agreement. Dotted lines denote random-patch results.}
\label{fig:trajectory}
\end{figure}

\paragraph{The distributed shift is coherent.}
The bottleneck positions outside the footprint receive change vectors with nearly identical directions and magnitudes. Let $O$ be these positions and $\Delta H_O$ their matrix of output changes. Define $g=|O|^{-1}(\Delta H_O)^\top\mathbf1$ as the mean change vector and $R_H=\Delta H_O-\mathbf1g^\top$ as the position-specific residual. Then
\begin{equation}
\Delta H_O=\mathbf1g^\top+R_H,\qquad
\norm{\Delta H_O}_F\geq\sqrt{|O|}\,\norm{g}_2-\norm{R_H}_F.
\label{eq:broadcast}
\end{equation}
Here $\mathbf1\in\mathbb R^{|O|}$ is an all-ones vector. The inequality follows from the reverse triangle inequality: when the residual is small, the common change contributes across all $|O|$ positions with aggregate magnitude proportional to $\sqrt{|O|}$. 
These results show how a small patch produces a coherent displacement over the full representation.

\subsection{The latent shift repeats across observations}\label{sec:stability}
\paragraph{Measuring transfer of the response.}
Section \ref{sec:propagation} explains how the patch reaches distant positions within one observation. 
Transfer across tasks also requires a similar response when the scene changes. For a fixed patch $p$, let $\Delta Z(x,p)=E(T_p(x))-E(x)$ be its final-latent change on observation $x$. 
We average the patch-induced latent changes in observations from 12 source-task states to form a fixed template \(\overline{\Delta Z}_{\mathrm{src}}(p)\), then compare it with changes in observations from 36 held-out states. The statistics in Table~\ref{tab:stability} use latent positions outside the projected patch footprint. Cross-task cosine similarity compares full final-latent displacement fields between tasks, while template cosine similarity compares each held-out field with the source template. Cosine similarities, relative residuals, and lower-bound ratios are reported as medians. Magnitude CV is the standard deviation of change magnitudes divided by their mean, computed over all 48 states.

\paragraph{One source template describes held-out scenes.}
The source template has a median cosine similarity of 0.981 with held-out changes, and the median residual $\norm{\Delta Z_i-\overline{\Delta Z}_{\mathrm{src}}}_1/\norm{\Delta Z_i}_1$ is 0.206 in Table~\ref{tab:stability}, where $i$ is the target held-out state. Clean representations from the same observations have a cross-task cosine similarity of 0.371, so the shared direction is specific to the patch-induced change. The random patch produces much weaker template agreement. Figure~\ref{fig:trajectory} shows that cross-observation alignment emerges early, while subsequent optimization mainly increases displacement on both source and held-out observations. Appendix~\ref{app:trajectory} reports the full trajectories, including the effective rank and frozen-template L1 bound during optimization.

\paragraph{The shared shift retains its magnitude.}
\ourwork{} maximizes L1 displacement, so the size of the common response matters alongside its direction. For a held-out state $i$, write its change as the source template plus a residual $e_i$. The reverse triangle inequality gives
\begin{equation}
\Delta Z_i=\overline{\Delta Z}_{\mathrm{src}}+e_i,\qquad\norm{\Delta Z_i}_1\geq\norm{\overline{\Delta Z}_{\mathrm{src}}}_1-\norm{e_i}_1.
\label{eq:templatebound}
\end{equation}
Using the measured residual for each held-out state, we compute the ratio of the lower bound in Equation~\eqref{eq:templatebound} to the observed L1 displacement. The median ratio is 0.833 (range 0.732--0.918). Thus, even after subtracting the observation-specific residual norm, the fixed source template yields a lower bound that retains much of the measured displacement.

\paragraph{Local features reproduce the broadcast.}
About 95.3\% of final-latent change energy outside the footprint lies in a shared spatial component. The direction of this component has a cross-task cosine similarity of about 0.997. To probe where this consistency begins, we replace keys and values within the patch footprint with source-task templates, retain a new image's clean context, and recompute attention. This K/V template analysis also includes a second patch initialization. The resulting broadcast closely matches the actual patched pass in direction and magnitude. Stable local keys and values thus help reproduce the distributed shift across observations.

\paragraph{Additional analysis.} We analyze how the latent shift changes actions in Appendix~\ref{app:behavior}.

%% file: sec/related.tex
\section{Related Work}

\paragraph{Transferable and localized adversarial attacks.}
Adversarial perturbations can alter neural-network predictions~\citep{ref_fgsm}. Cross-model transfer has been studied using ensembles~\citep{ref_delving}, momentum~\citep{ref_momentum}, and input diversity~\citep{ref_diversity}. Universal perturbations reuse a single perturbation across images~\citep{ref_uap}, with data-independent constructions such as Fast Feature Fool~\citep{ref_fff}. Feature-level objectives have been explored through feature disruption~\citep{ref_fda}, feature importance~\citep{ref_fia}, and neuron attribution~\citep{ref_naa}. Localized attacks include LaVAN~\citep{ref_lavan} and attention-aware Patch-Fool~\citep{ref_patchfool}; physical attacks include robust object perturbations~\citep{ref_rp2} and camera stickers~\citep{ref_stickers}. PhotoGuard also studies encoder-level attacks on diffusion-based image editing~\citep{ref_photoguard}. In sequential decision-making, adversarial policies can disrupt reinforcement-learning agents through an opponent's behavior~\citep{ref_advpolicies}.

\paragraph{Attack on vision-language-action policies.}
Vision-language-action (VLA) policies such as OpenVLA~\citep{ref_openvla} and $\pi_0$~\citep{ref_pi0} map visual and language inputs to robot actions. Action-based attacks on VLA policies optimize action differences, motion directions, target actions, or action freezing~\citep{r8,r9,r10}. Their gradient-based objectives require access through the corresponding action generation interfaces. Representation-based methods attack image--text alignment, feature displacement, projection features, attention relationships, or visual proprioception~\citep{r3,r4,r11,r12,r13}. Clean--perturbed representation distances already appear in prior attacks; here, we study their effectiveness for fixed local attacks on a public generative visual encoder and the mechanism of transfer. In the WAM pathway studied here, image--text interaction occurs downstream of the public VAE, so the corresponding alignment and attention objectives fall outside the default encoder-only access. We therefore define the objective directly on the encoder output and evaluate its effects on control through closed-loop experiments and mechanism interventions.

\paragraph{Adversarial threats to world models and world action models.}
BadWorld~\citep{r18} evaluates attacks on generated videos, without testing closed-loop robotic tasks. BadWAM~\citep{r5} perturbs the full visual input through online target-model queries at each replanning step. Attacking the Trusted Imagination~\citep{r17} optimizes observation perturbations using end-to-end gradients through the encoder and world model, corrupting future predictions used by downstream planning or safety checks. JailWAM~\citep{r16} induces unsafe actions and requires attacker control over language instructions. Unlike these WAM threats, \ourwork{} achieves closed-loop attacks across tasks and WAM architectures with a fixed local patch optimized solely through a public visual encoder, without using target-policy action outputs or future predictions.

%% file: sec/conclusion.tex
\section{Conclusion}
\ourwork{} optimizes a fixed local patch using only a public visual encoder and six observations from one task, yet produces strong closed-loop interference across two complete benchmarks and different downstream action architectures. Mechanism analysis reveals that interactions between attention-weight and value changes broadcast local interference to distant representations, while optimized perturbation responses remain highly consistent across task observations. Behavioral analysis connects this representation vulnerability to grasping and pose interference, and separates the immediate effects of sustained visual input from state consequences accumulated through execution. These results show that WAM robustness must cover the full path from local visual input through representation, generation, and control.